# 基于自步学习的对称非负矩阵分解算法

王 雷[1]，杜 亮[1,2]，周 芃[3]，吴 鹏[1,2]

(1.山西大学 计算机与信息技术学院 山西 太原 030006; 2.山西大学 大数据科学与产业研究院 山西 太原 030006; 3.安徽大学 计算机科学与技术学院 安徽 合肥 230601)

摘要: 提出一种基于自步学习的对称非负矩阵分解算法,通过误差驱动的方式使模型更好地区分正常样本与异常样本,进而提高模型的聚类性能。该方法为所有样本赋予了一个可以衡量其难易程度的权重变量,并采用硬加权与软加权两种策略分别对此变量进行约束以保证模型的合理性。在图像、文本等多个数据集上进行聚类分析,实验结果表明了所提算法的有效性。

关键词: 无监督学习; 对称非负矩阵分解; 误差驱动; 自步学习; 聚类

# Symmetry Nonnegative Matrix Factorization Algorithm Based on Self-paced Learning

WANG Lei[1], DU Liang[1,2], ZHOU Peng[3], WU Peng[1,2]

(1.*College of Computer and Information Technology*, *Shanxi University*, *Taiyuan* 030006, *China*; 2.*Institute of Big Data Science and Industry*, *Shanxi University*, *Taiyuan* 030006, *China*; 3.*College of Computer Science and Technology*, *Anhui University*, *Hefei* 230601, *China*)

**Abstract**: A symmetric nonnegative matrix factorization algorithm based on self-paced learning was proposed to improve the clustering performance of the model. It could make the model better distinguish normal samples from abnormal samples in an error-driven way. A weight variable that could measure the degree of difficulty to all samples was assigned in this method, and the variable was constrained by adopting both hard-weighting and soft-weighting strategies to ensure the rationality of the model. Cluster analysis was carried out on multiple data sets such as images and texts, and the experimental results showed the effectiveness of the proposed algorithm.

**Key words**: unsupervised learning; symmetry nonnegative matrix factorization; error-driven; self-paced learning; clustering

## 0 引言

聚类是机器学习常见的任务之一[1-2],非负矩阵分解(NMF)[3]和 K-means[4]都是其重要且经典的算法。但 NMF 存在一定的局限性: 当数据嵌入非线性流形时,NMF 不再适用于将其解释为一个原始数据矩阵分解为一个基矩阵与其加权和矩阵。因此,产生了一种 NMF 的变体——对称非负矩阵分解(SNMF)[5]。一般 SNMF 的输入为样本间的关系矩阵,虽然 SNMF 相对于 NMF 有了更好的表示能力,但是增加了计算的复杂度。随着对 SNMF 问题的研

收稿日期: 2021-06-01

基金项目: 国家自然科学基金项目(61502289,61806003);山西省重点研发项目(201803D31199);山西省自然科学基金项目(201801D221163,201801D221173)。

第一作者: 王雷(1996—),男,硕士研究生,主要从事聚类算法研究,E-mail: 575264909@qq.com。

通信作者: 杜亮(1985—),男,副教授,主要从事聚类算法、特征选择和鲁棒研究,E-mail: duliang@sxu.edu.cn。

究,出现了很多有效的求解算法,例如带有 $\alpha$ 和 $\beta$ 因子的 MUR 算法[6];投影梯度下降算法[7];块坐标下降算法[8];贪婪梯度下降算法[9];交替非负最小二乘算法[10]等。但是,SNMF 在构造样本的关系矩阵时,会不可避免地出现错误判断的情况。

为解决上述问题,文献[11]引入自步学习框架。自步学习的思想源于课程学习[12],其核心思想是逐步增加学习的难度来改善模型的训练能力。自步学习有着广泛的应用场景,例如自步矩阵分解[13];自步聚类集成[14];基于自步学习正则的无监督特征选择[15]等。本文提出一种基于自步学习的对称非负矩阵分解(SPSNMF)算法,利用自步学习的特性,根据样本损失的大小赋予样本一个可衡量其难易程度的权重变量,以更好地区分正常样本和异常样本,以此来增加模型的稳定性,进而提高聚类任务的准确性。在不同数据集上与多种聚类算法进行对比,实验结果表明了所提算法的有效性。

# 1 相关工作

## 1.1 对称非负矩阵分解

对称非负矩阵分解问题是给出一个相似性矩阵 $X\in\mathbf{R}^{n\times n}$,其表示一个大小为 $n\times n$ 的对称矩阵,$U\in\mathbf{R}^{n\times k}$为分解得到的低秩矩阵。对称非负矩阵分解的一般问题可以描述为

$$\min_{U\geqslant 0}\|X-UU^{\mathrm{T}}\|_F^2, \tag{1}$$

其中:$U\geqslant 0$ 与非对称矩阵分解约束一致。而此时对于模型的求解,从一个二次规划问题转变为一个四次优化问题,复杂度提升。

## 1.2 自步学习

在聚类任务中,由于数据点的差异性和离群点的存在,在模型训练初期会产生巨大的阻碍,使得模型训练出现偏差,通过加入自步学习框架可以缓解此问题的发生。含有自步学习框架的优化模型可以表示为

$$\min_{w\in[0,1]^n}\sum_{i=1}^{n}w_iL(\theta;x_i,y_i)+f(w;\lambda), \tag{2}$$

其中:$L(\theta;x_i,y_i)$ 为模型的损失函数,$\theta$ 为决策模型中产生的参数;$w=[w_1,w_2,\cdots,w_n]^{\mathrm{T}}$ 为用来衡量样本难易程度的权重变量;$\lambda$ 为控制自步函数学习步长的参数。自步函数大体可分为以下两类。

一类是被称为硬加权的自步函数。其一般形式为 $f(w;\lambda)=-\frac{1}{\lambda}\sum_{i=1}^{n}w_i$。$w$ 的更新规则为

$$w_i^*(\lambda;l_i)=\begin{cases}1, & \sum_j l_{ij}\leqslant\frac{1}{\lambda},\\ 0, & \sum_j l_{ij}>\frac{1}{\lambda},\end{cases} \tag{3}$$

其中:$w_i^*(\lambda,l_i)=\min_{w_i\in\{0,1\}}w_il_i+f(w;\lambda)$。式(3)表明,当样本的损失不大于某一个阈值 $1/\lambda$ 时,判断其为简单样本,赋予其权重 1 并纳入训练模型;反之,当样本的损失大于阈值 $1/\lambda$ 时,则赋予其权重 0 且不加入训练模型。

另一类是被称为软加权的自步函数。文献[16]给出了软自步正则的多种形态,本文的混合软自步函数形式为 $f(w;\lambda,\lambda')=-\zeta(\sum_{i=1}^{n}\log(w_i+\zeta\lambda))$。$w$ 的更新规则为

$$w_i^*(\lambda',\lambda;l_i)=\begin{cases}1, & \sum_j l_{ij}\leqslant\frac{1}{\lambda'},\\ 0, & \sum_j l_{ij}\geqslant\frac{1}{\lambda},\\ \frac{\zeta}{l_i}-\lambda\zeta, & 其他,\end{cases} \tag{4}$$

其中:$\zeta=\frac{1}{\lambda'-\lambda}$,$\lambda'>\lambda>0$。

# 2 基于自步学习的对称非负矩阵分解算法

## 2.1 模型建立

本文模型可以描述为

$$F(U,w)=\min_{U,w}\frac{1}{2}\|(X-UU^{\mathrm{T}})\otimes diag(\sqrt{w})\|_F^2+f(w;\lambda),\mathrm{s.t.}\ U\geqslant 0,w\geqslant 0, \tag{5}$$

其中:$\otimes$表示点积;$diag(\cdot)$ 表示对角矩阵;$w$ 为融入自步学习框架后赋予样本的权重变量,用来衡量每个样本的难易程度;$f(w;\lambda)$ 项用来限定权重,确保模型的合理性。

## 2.2 模型求解

本文算法的求解采用交替迭代的求解方式。对于 $U$、$V$ 的求解采用文献[17]中的求解思路,即采用非对称形式的 SNMF 模型求解,模型可以描述为

$$\min_{U\geqslant 0,V\geqslant 0}\frac{1}{2}\|X-UV^{\mathrm{T}}\|_F^2+\frac{\theta}{2}\|U-V\|_F^2, \tag{6}$$

其中:$\theta>0$ 为平衡参数。当 $\theta$ 足够大时,式(6)中的任何有界临界点都能满足 $U=V$,并能取到与式(1)中一样的关键点。因此,模型可以重写为

$$F(U,V,w)=\min_{U,V,w}\frac{1}{2}\|(X-UV^{\mathrm{T}})\otimes diag(\sqrt{w})\|_F^2+$$

$$\frac{\theta}{2}\|U-V\|_F^2+f(w;\lambda),\text{s.t. } U\geqslant 0,V\geqslant 0,w\geqslant 0。\tag{7}$$

对于式(7),$U$、$V$、$w$ 采取交替更新的方式来进行求解,具体如下。

1) 固定 $w$

当更新 $U$ 和 $V$ 时,模型可以描述为

$$F(U)=\min_U \frac{1}{2}\|(X-UV)\otimes diag(\sqrt{w})\|_F^2+\frac{\theta}{2}\|U-V\|_F^2,\text{s.t. } U\geqslant 0。\tag{8}$$

对于式(8),将 $U$、$V$ 表示为列向量($u_1,\cdots,u_k,v_1,\cdots,v_k$)。在更新 $U$ 或 $V$ 时,采用交替更新 $u_i$ 及 $v_i$($i=1,\cdots,k$),直至达到模型收敛为止,此时 $X_i$ 可以通过 $X_i=X-\sum_{i\neq j}u_jv_j^{\mathrm{T}}$ 计算,即每次的 $X_i$ 都是由除了 $u_i$ 和 $v_i$ 以外的其他元素决定的,故式(8)可以重写为

$$F(u_i^t)=\arg\min_{u_i}\frac{1}{2}w_i\|(X_i^t-u_i(v_i^{t-1})^{\mathrm{T}})\|_F^2+\frac{\theta}{2}\|u_i-v_i^{t-1}\|_F^2,\text{s.t. } u_i^t\geqslant 0,\tag{9}$$

其中:$t$ 表示迭代次数。当固定 $v_i$ 和 $w_i$ 时,式(9)是一个变量为 $u_i$ 的二次方程,对其求导并令其为 0,可得 $u_i$ 的更新公式为

$$(u_i^t)^*=\max\left(\frac{\sum_{j=1}^{n}(X_{ij}^tw_j+\theta)v_{ij}^{t-1}}{\sum_{j=1}^{n}(v_{ij}^{t-1}v_{ij}^{t-1}w_j)+\theta},0\right)。\tag{10}$$

同理,当固定 $u_i$ 和 $w_i$ 时,$v_i$ 的更新公式为

$$(v_i^t)^*=\max\left(\frac{\sum_{j=1}^{n}(X_{ij}^tw_j+\theta)u_{ij}^t}{\sum_{j=1}^{n}(u_{ij}^tu_{ij}^tw_j)+\theta},0\right)。\tag{11}$$

因为 $u_i$ 和 $v_i$ 有着非负约束,所以需要给出一个下确界 0。

2) 更新 $w$

当固定 $u_i$ 和 $v_i$ 时,模型可以描述为

$$F(w)=\arg\min_{w_j}\sum_{j=1}^{n}w_jl_{ij}+f(w,l)。\tag{12}$$

式(12)是一个变量为 $w_i$ 的凸函数,$w_i$ 的两种自步函数策略的更新规则分别对应式(3)、(4)。

根据上述描述,本文 SPSNMF 算法的具体步骤如下。

**算法 1　自步对称非负矩阵分解算法**

输入: 数据集通过 $K$ 近邻算法构造相似度矩阵 $X$;聚类数目 $k$。

输出: 分解矩阵 $U$。

1: 初始化 $U_0$ 和 $V_0$ 并计算平衡因子 $\theta$;迭代次数 $t=0$;

2: 通过计算所有样本的损失函数,选取中位数作为初始自步参数 $\lambda$ 的值;

3: while not convergence

4: 　通过式(3)、(4)更新 $w$;

5: 　for $i=1$: $k$

6: 　　通过公式 $X_i=X-\sum_{i\neq j}u_jv_j^{\mathrm{T}}$ 计算 $X_i$;

7: 　　通过式(10)更新 $U_i$;

8: 　　通过式(11)更新 $V_i$;

9: 　end for

10: 通过式(7)计算目标函数 $F_t$;

11: 当 $\frac{F_{t-1}-F_t}{F_{t-1}}<10^{-6}$ 时,判定收敛,终止迭代;

12: end while

13: return $U$

### 2.3　模型分析

**2.3.1　参数分析**　式(9)的第一项为 $F$ 范数衡量的误差损失,第二项为引入的正则化因子,目的是迫使 $U$ 和 $V$ 相等,其中参数 $\theta$ 的目的是为了平衡第一项和第二项的正则项。当 $\theta$ 足够大时,模型优化会迫使 $\|U-V\|_F^2$ 趋于 0,从而使 $U$ 趋向等于 $V$。根据文献[17]中定理 2,$\theta$ 的选取范围为

$$\theta>\frac{1}{2}(\|X\|_2+\|X-U_0U_0^{\mathrm{T}}\|_F-\sigma_n(X)),\tag{13}$$

其中:$\|X\|_2$ 表示 $X$ 最大的奇异值;$\sigma_i(X)$ 表示 $X$ 的第 $i$ 个最大奇异值;$U_0$ 表示 $U$ 的初始选取值。

**2.3.2　时间复杂度分析**　当更新 $U$ 时,时间复杂度为 $O(n^2k+nk^2)$,更新 $V$ 的时间复杂度与 $U$ 类似。当更新 $w$ 时,复杂度为 $O(n)$。因此,更新所需的时间复杂度为 $O(n^2k+nk^2)$,由于 $k\ll n$,故复杂度为 $O(n^2k)$。假设所有样本全部纳入训练时共进行了 $P$ 轮(这里的 $P\ll n$),算法在 $T$ 次迭代后达到收敛条件,故本文算法的复杂度为 $O(TPn^2k)$。

## 3　实验与结果分析

对本文所提出的硬加权的自步学习框架(HSPSNMF)和软加权的自步学习框架(SSPSNMF)进行性能评估,将实验参数进行如下设置:平衡因子 $\theta$ 通过式(13)计算,并向上取整;$\lambda$ 的第 1 次选取为

[0.5,1],随后每次增加10%的样本;设置自步权重在每迭代10次时更新1次,直至所有的样本被纳入训练。为了对比实验的公平性,对每个数据集进行10次实验,取平均值。

## 3.1 对比方法和数据集

为了衡量本文SPSNMF算法的性能,将其与其他几种密切相关的NMF方法和经典聚类方法进行对比分析。对比方法包括NMF[3];Ncut[18];$\alpha$-SNMF[6];CIMNMF[19];$L_{2,1}$NMF[20];SPLNMF[21];SymHALS[17];CauchyNMF[22]。在对比实验中,将Ncut算法的参数设置为$\delta \in [2^{-3},2^{3}]$,$k \in [5,10,15]$;将$\alpha$-SNMF算法的$\alpha$设置为0.99;CauchyNMF算法采用Nesterov最优梯度法和原始数据自适应加权。数据集的描述如表1所示。

表1 数据集的描述

**Table 1** Description of the data sets

| 数据集 | 样本个数 | 特征个数 | 类别个数 |
|---|---|---|---|
| BRAIN | 42 | 5 597 | 5 |
| ALLAML | 72 | 7 129 | 2 |
| GLIOMA | 50 | 4 434 | 4 |
| LYMPHOMA | 96 | 4 026 | 9 |
| JAFFE | 231 | 676 | 10 |
| UMIST | 575 | 644 | 20 |
| FBIS | 2 463 | 2 000 | 17 |
| TDT2 | 9 394 | 36 771 | 30 |

## 3.2 收敛性与参数灵敏度分析

以JAFFE数据集为例,将本文HSPSNMF算法与SymHALS算法进行聚类质量和收敛情况对比分析,结果如图1和图2所示。本文HSPSNMF和SSPSNMF算法在JAFFE数据集上添加不同比例样本对聚类精度的影响如图3和图4所示。

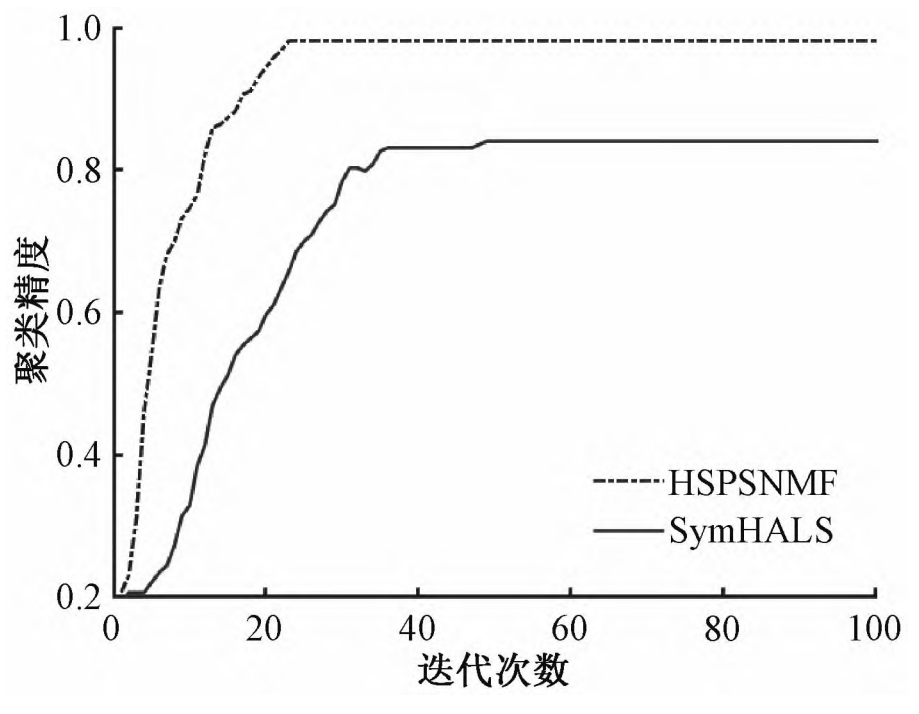

图1 HSPSNMF和SymHALS算法在JAFFE数据集上的聚类质量情况

**Figure 1** Clustering quality of HSPSNMF and SymHALS algorithms on JAFFE data set

图1和图2显示,与SymHALS算法对比,

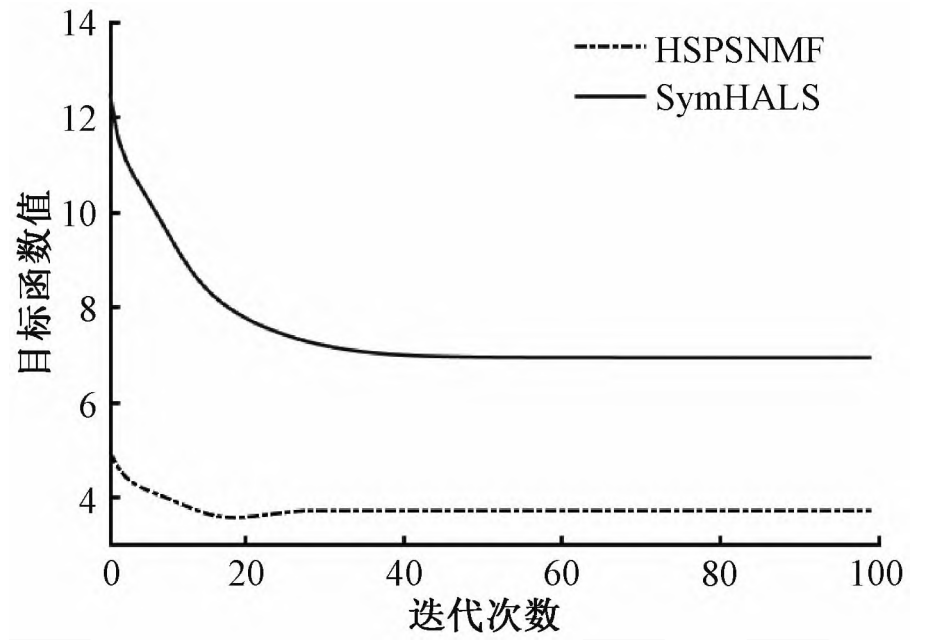

图2 HSPSNMF和SymHALS算法在JAFFE数据集上的目标函数收敛情况

**Figure 2** Objective function convergence of HSPSNMF and SymHALS algorithms on JAFFE data set

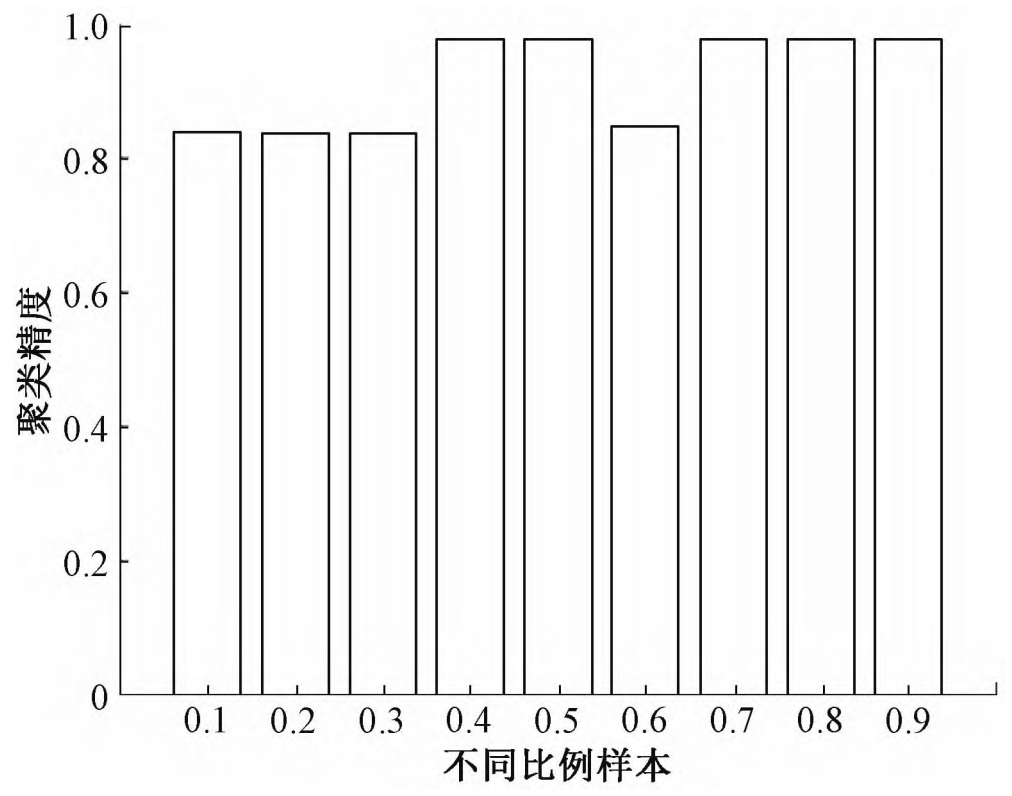

图3 HSPSNMF算法在JAFFE数据集上添加不同比例样本对聚类精度的影响

**Figure 3** Influence of adding different proportion samples to JAFFE data set by HSPSNMF algorithm on clustering accuracy

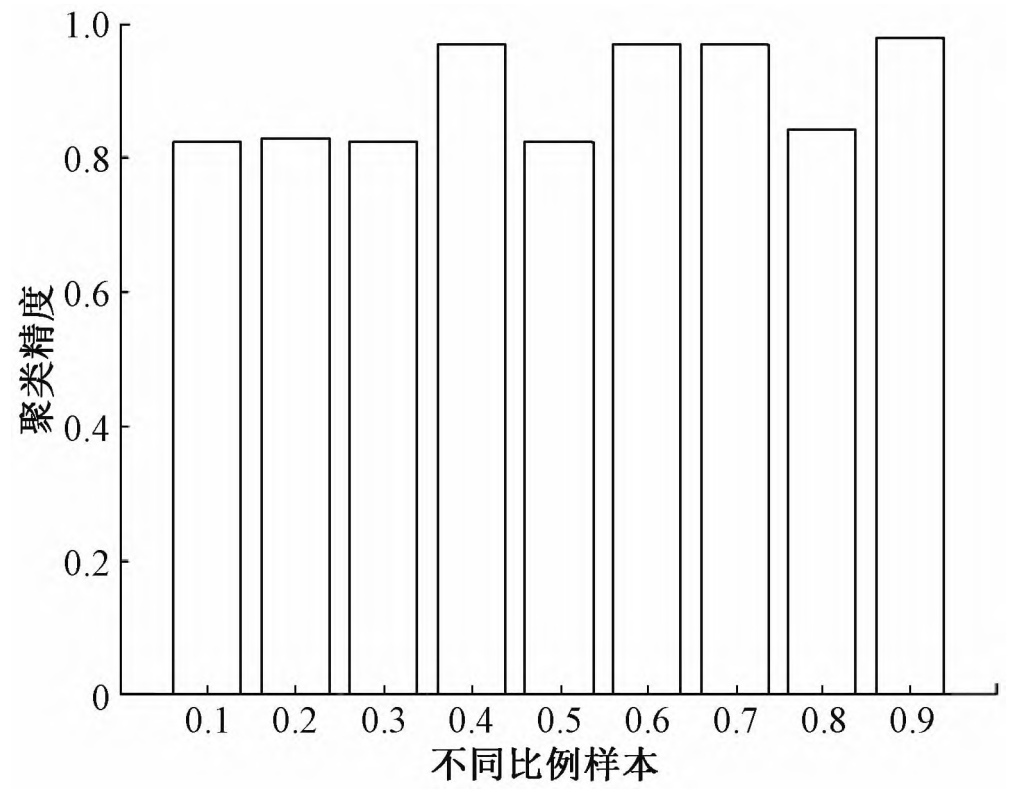

图4 SSPSNMF算法在JAFFE数据集上添加不同比例样本对聚类精度的影响

**Figure 4** Influence of adding different proportion samples to JAFFE data set by SSPSNMF algorithm on clustering accuracy

HSPSNMF算法在迭代20次左右即可达到收敛。图3和图4显示,SPSNMF算法中的超参数在初始选取样本比例大于0.4时效果更好。

### 3.3　评价指标及实验结果

采用的聚类评价指标为聚类精度(*ACC*)、归一化互信息(*NMI*)、调兰德指数(*ARI*),聚类性能结果分别汇总于表 2~4。可以看出,本文方法适用于多个数据集,并且在聚类性能上优于对比算法。

表 2　*ACC* 实验结果对比

Table 2　Comparison of *ACC* experimental results

| 数据集 | *ACC* | | | | | | | | | |
|---|---|---|---|---|---|---|---|---|---|---|
| | NMF | Ncut | α-SNMF | CIMNMF | $L_{2,1}$NMF | SPLNMF | CauchyNMF | SymHALS | HSPSNMF | SSPSNMF |
| BRAIN | 0.459 5 | 0.583 3 | 0.742 9 | 0.507 1 | 0.397 6 | 0.261 9 | 0.669 0 | 0.750 0 | 0.785 7 | 0.688 1 |
| ALLAML | 0.655 6 | 0.638 9 | 0.634 7 | 0.663 9 | 0.641 7 | 0.597 2 | 0.622 2 | 0.616 7 | 0.708 3 | 0.702 8 |
| GLIOMA | 0.424 0 | 0.558 0 | 0.584 0 | 0.452 0 | 0.422 0 | 0.472 0 | 0.440 0 | 0.604 0 | 0.616 0 | 0.696 0 |
| LYMPHOMA | 0.372 9 | 0.395 8 | 0.512 5 | 0.415 6 | 0.358 3 | 0.402 1 | 0.492 7 | 0.591 7 | 0.619 8 | 0.583 3 |
| JAFFE | 0.768 1 | 0.907 0 | 0.689 7 | 0.916 9 | 0.799 1 | 0.432 4 | 0.791 1 | 0.889 7 | 0.874 6 | 0.928 2 |
| UMIST | 0.412 5 | 0.088 7 | 0.457 6 | 0.376 3 | 0.412 5 | 0.309 9 | 0.132 0 | 0.499 1 | 0.511 0 | 0.519 0 |
| FBIS | 0.356 9 | 0.248 4 | 0.344 1 | 0.291 6 | 0.366 1 | 0.300 7 | 0.376 7 | 0.339 2 | 0.376 7 | 0.373 1 |
| TDT2 | 0.432 2 | 0.129 2 | 0.515 3 | 0.174 8 | 0.465 0 | 0.208 4 | 0.075 3 | 0.646 6 | 0.692 4 | 0.690 3 |
| 平均值 | 0.485 2 | 0.443 7 | 0.560 1 | 0.474 8 | 0.482 8 | 0.373 1 | 0.449 9 | 0.617 1 | 0.648 1 | 0.647 6 |

表 3　*NMI* 实验结果对比

Table 3　Comparison of *NMI* experimental results

| 数据集 | *NMI* | | | | | | | | | |
|---|---|---|---|---|---|---|---|---|---|---|
| | NMF | Ncut | α-SNMF | CIMNMF | $L_{2,1}$NMF | SPLNMF | CauchyNMF | SymHALS | HSPSNMF | SSPSNMF |
| BRAIN | 0.307 3 | 0.552 8 | 0.644 0 | 0.370 0 | 0.211 7 | 0.098 6 | 0.534 2 | 0.679 0 | 0.713 4 | 0.607 5 |
| ALLAML | 0.081 1 | 0.009 3 | 0.075 3 | 0.074 1 | 0.063 4 | 0.024 8 | 0.094 2 | 0.076 7 | 0.112 8 | 0.107 6 |
| GLIOMA | 0.166 6 | 0.353 3 | 0.497 7 | 0.188 7 | 0.174 6 | 0.187 6 | 0.196 1 | 0.508 1 | 0.443 0 | 0.533 2 |
| LYMPHOMA | 0.342 5 | 0.038 5 | 0.572 1 | 0.396 3 | 0.352 9 | 0.042 0 | 0.504 0 | 0.619 3 | 0.639 6 | 0.596 7 |
| JAFFE | 0.810 4 | 0.929 9 | 0.792 1 | 0.931 9 | 0.840 8 | 0.386 9 | 0.840 9 | 0.934 2 | 0.929 1 | 0.954 1 |
| UMIST | 0.593 6 | 0.049 3 | 0.613 4 | 0.547 0 | 0.591 0 | 0.424 2 | 0.131 1 | 0.705 3 | 0.722 1 | 0.722 3 |
| FBIS | 0.321 8 | 0.045 7 | 0.342 0 | 0.242 3 | 0.326 8 | 0.131 3 | 0.386 7 | 0.357 1 | 0.386 7 | 0.392 8 |
| TDT2 | 0.468 9 | 0.077 9 | 0.562 2 | 0.140 8 | 0.490 1 | 0.162 6 | 0.008 5 | 0.717 6 | 0.757 1 | 0.758 6 |
| 平均值 | 0.386 5 | 0.257 1 | 0.512 4 | 0.361 4 | 0.381 4 | 0.182 4 | 0.337 0 | 0.574 7 | 0.588 0 | 0.584 1 |

表 4　*ARI* 实验结果对比

Table 4　Comparison of *ARI* experimental results

| 数据集 | *ARI* | | | | | | | | | |
|---|---|---|---|---|---|---|---|---|---|---|
| | NMF | Ncut | α-SNMF | CIMNMF | $L_{2,1}$NMF | SPLNMF | CauchyNMF | SymHALS | HSPSNMF | SSPSNMF |
| BRAIN | 0.156 6 | 0.384 3 | 0.549 6 | 0.233 2 | 0.080 8 | 0.003 4 | 0.415 5 | 0.562 8 | 0.604 9 | 0.500 4 |
| ALLAML | 0.088 5 | −0.012 9 | 0.081 6 | 0.107 6 | 0.061 0 | −0.028 0 | 0.066 7 | 0.063 8 | 0.161 0 | 0.152 2 |
| GLIOMA | 0.076 3 | 0.210 0 | 0.359 7 | 0.080 3 | 0.079 7 | 0.090 4 | 0.098 5 | 0.380 3 | 0.312 8 | 0.414 9 |
| LYMPHOMA | 0.139 3 | −0.091 7 | 0.262 9 | 0.159 5 | 0.115 4 | −0.084 1 | 0.232 8 | 0.331 7 | 0.367 7 | 0.316 5 |
| JAFFE | 0.693 4 | 0.882 7 | 0.604 5 | 0.882 6 | 0.733 0 | 0.094 7 | 0.728 2 | 0.873 4 | 0.862 0 | 0.914 1 |
| UMIST | 0.289 0 | 0.000 4 | 0.302 4 | 0.247 6 | 0.288 7 | 0.112 4 | 0.002 3 | 0.407 1 | 0.425 1 | 0.431 8 |
| FBIS | 0.203 4 | 0.038 2 | 0.165 6 | 0.168 3 | 0.161 5 | 0.032 8 | 0.200 0 | 0.160 3 | 0.200 0 | 0.200 8 |
| TDT2 | 0.111 1 | 0.015 6 | 0.251 5 | 0.032 5 | 0.116 1 | 0.007 2 | 0.000 9 | 0.548 5 | 0.610 4 | 0.614 6 |
| 平均值 | 0.219 7 | 0.178 3 | 0.322 2 | 0.239 0 | 0.204 5 | 0.028 6 | 0.218 1 | 0.416 0 | 0.443 0 | 0.443 2 |

## 4　小结

本文提出一种基于自步学习的对称非负矩阵分解(SPSNMF)算法,探索了 SNMF 在加入自步学习框架以后,对其聚类性能的改变以及收敛性。SPSNMF 算法具有由简入难地逐步加入样本学习到模型的特性,在很大程度上改善了模型的稳定性和

SNMF 中相似性矩阵构造时出现错误判断的情况，在聚类性能上相较于其他方法更加稳定。但本文提出的 SPSNMF 算法存在以下缺陷：在加入自步学习框架后，模型的迭代求解次数增加，进而大大增加了模型的求解时间；对于自步学习中的两种加权策略，未挖掘出其在理论上的可行性以及在不同场景中的适用性。在未来的工作中将进一步探索更高效的 SPSNMF 求解算法，以及算法在不同类型数据集(例如平衡与非平衡数据)中的应用。

## 参考文献：